# Anomaly Detection with Adversarial Dual Autoencoders

**Vu Ha Son**[1], **Ueta Daisuke**[2], **Hashimoto Kiyoshi**[2],
**Maeno Kazuki**[3], **Sugiri Pranata**[1], **Sheng Mei Shen**[1]
[1]Panasonic R&D Center Singapore
`{hason.vu,sugiri.pranata,shengmei.shen}@sg.panasonic.com`
[2]Panasonic CNS – Innovation Center
[3]Panasonic CETDC
`{ueta.daisuke,hashimoto.kiyoshi,maeno.kazuki}@jp.panasonic.com`

## Abstract

Semi-supervised and unsupervised Generative Adversarial Networks (GAN)-based methods have been gaining popularity in anomaly detection task recently. However, GAN training is somewhat challenging and unstable. Inspired from previous work in GAN-based image generation, we introduce a GAN-based anomaly detection framework – Adversarial Dual Autoencoders (ADAE) - consists of two autoencoders as generator and discriminator to increase training stability. We also employ discriminator reconstruction error as anomaly score for better detection performance. Experiments across different datasets of varying complexity show strong evidence of a robust model that can be used in different scenarios, one of which is brain tumor detection.

## 1   Introduction

The task of anomaly detection is informally defined as follows: given the set of normal behaviors, one must detect whether incoming input exhibits any irregularity. In anomaly detection, semi-supervised and unsupervised approaches have been dominant recently, as the weakness of supervised approaches is that they require monumental effort in labeling data. On the contrary, semi-supervised and unsupervised methods do not require many data labeling, making them desirable, especially for rare/unseen anomalous cases.

Out of the common methods for semi and unsupervised anomaly detection such as variational autoencoder (VAE), autoencoder (AE) and GAN, GAN-based methods are among the most popular choices. However, training of GAN is not always easy, given problems such as mode collapse and non-convergence, often attributed to the imbalance between the generator and discriminator. One way to mitigate these problems is to have generator and discriminator of the same type, i.e. autoencoders, as proposed in EBGAN [1] and BEGAN [2] for generating realistic images.

Inspired by this, we propose a semi-supervised GAN-based method for anomaly detection task such that both generator and discriminator are made up of autoencoders, called Adversarial Dual Autoencoders (ADAE). Anomalies during testing phase are detected using discriminator pixel-wise reconstruction error. This method is tested on multiple datasets such as MNIST, CIFAR-10 and Multimodal Brain Tumor Segmentation (BRATS) 2017 dataset and is found to achieve state-of-the-art results.

The outline of the paper is organized as follows. Section 2 reviews the related GAN-based methods for anomaly detection. Section 3 emphasizes on the motivation and details of ADAE. Section 4 and 5 describe the experimental setup and demonstrate the state-of-the-art performance of ADAE on various datasets.

## 2   Related Work

### 2.1   GAN-based anomaly detection methods

For anomaly detection task, it is also important to map the image space to latent space ($X \rightarrow Z$) which cannot be obtained from the usual GAN training. Thus, an encoder is usually used for such purpose.

GANomaly [3] utilizes an encoder($G_E$)-decoder($G_D$)-encoder($E$) pipeline architecture, whereby $G_E$ learns to map $X \rightarrow Z$. The model learns to minimize the latent space representation reconstruction error in the two encoders for normal data. Anomaly score $A(\hat{x})$ of test input $\hat{x}$ is calculated from latent space reconstruction error:

$$A(\hat{x}) = \|G_E(\hat{x}) - E(G(\hat{x}))\|_1$$

Where $G_E(\hat{x})$ is the latent representation of inputs in generator encoder and $E(G(\hat{x}))$ is the generated image's latent representation in the second encoder. Anomalous samples would cause the model to fail to reconstruct latent representation in the second encoder, result in high anomaly scores.

EGBAD [4], on the other hand, makes use Bidirectional GAN (BiGAN) [5] to learn an encoder which is the inverse of the generator to map image space to latent space, i.e. $E = G^{-1}$. Anomaly score $A(\hat{x})$ is then calculated with a combination of reconstruction loss $L_G$ and discriminator-based loss $L_D$:

$$A(\hat{x}) = \alpha L_G(\hat{x}) + (1 - \alpha)L_D(\hat{x})$$

Similarly, IGMM-GAN [6] also draws inspiration from BiGAN to learn the latent space representation of multimodal normal data. It further explores the latent space representation, using IGMM to learn the clustering in latent space. Anomaly scores of each test samples are the Mahalanobis distance to the nearest cluster, derived from the means and variances of different clusters. Anomalies are thought to be points further away from the learned clusters.

## 2.2 Semi-supervised/Unsupervised Brain Tumor Detection

In the domain of brain tumor detection/segmentation, semi-supervised and unsupervised methods have also achieved encouraging performance, without the need for labeled data. The main theme of semi-supervised and unsupervised tumor segmentation (pixel-wise detection) methods is that the model would fail to reconstruct the tumor, having only learnt normal brain data. The residual image between test inputs and reconstructed image would then highlight the tumor. Baur et al. proposed AnoVAEGAN [7] combines VAE and GAN to form an end-to-end model. Chen and Konukoglu [8] used Adversarial Autoencoder (AAE) along with consistency constrained in the latent space to enforce the reconstruction of brain image except the tumor.

In terms of tumor detection (slice-wise detection), Zimmerer et al. [9] made use of both context-encoding and VAE to form ceVAE, which provides a better reconstruction error and model-internal variations. Anomaly score is the estimated probability of testing sample, derived from evaluating the ELBO of the test inputs.

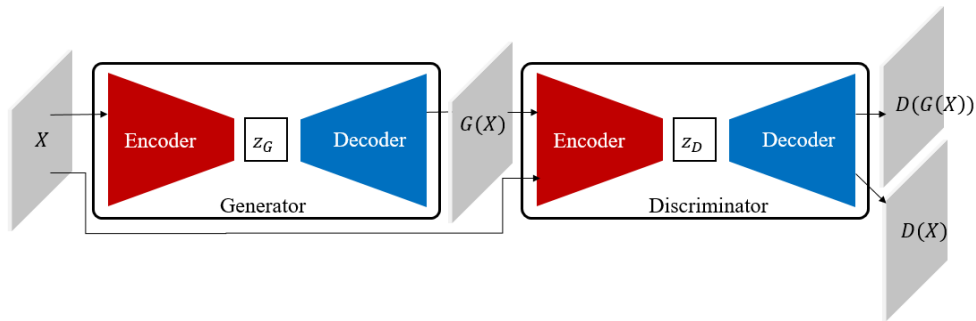

**Figure 1:** Architecture of ADAE with two autoencoders as generator and discriminator.

## 3 Proposed Method: ADAE

GAN was first introduced by Goodfellow et al. [10]. Its main idea involves a generator G and a discriminator D having two competing objectives: the generator attempts to produce realistic images that likely belong to the data distribution of interest $X$, while the discriminator tries to distinguish between $X$ and generated images $G(X)$. In theory, the training converges when the generator can "fool" the discriminator by generating images $G(X)$ that match distribution of $X$. The general training objective in practice of GAN is given as:

$$\min_G \max_D V(D, G) = \mathbb{E}_{x \sim p_{data}(x)}[\log D(x)] - \mathbb{E}_{z \sim p_z(z)}[\log D(G(z))]$$

It is common to have a binary classifier as the discriminator, explicitly trained to decide whether an image is real or fake. However, this gives rise to an imbalance of capability between the two networks which results in unstable training and hinders the ability of GAN. While there are ways to make training become more stable by putting restraints on the discriminator [11] [12], others suggest having an auto-encoder as discriminator [1] [2] to achieve a better balance. This change in structure, coupled with appropriate training objectives, can help to achieve images of good quality.

Thus, we propose ADAE, a network architecture with two sub-networks made up of autoencoders. In order to fulfil the typical role of the respective GAN sub-networks, the adversarial training objective is defined as the pixel-wise error between the reconstructed image of data $X$ through $G$ and of generated image $G(X)$ through $D$:

$$\|G(X) - D(G(X))\|_1$$

The discriminator is trained to maximize this error, which causes it to fail to reconstruct the inputs if they are thought to belong to generated distribution. In essence, the discriminator attempts to separate the real and generated data distributions. On the other hand, the generator is trained to minimize this error, forces it to produce images in the real data distribution so as to "fool" the discriminator.

The discriminator training objective is to reconstruct faithfully the real inputs while fail to do so for generated input, as shown below:

$$\mathcal{L}_D = \|X - D(X)\|_1 - \|G(X) - D(G(X))\|_1$$

The generator, besides trying to reconstruct the real input, would also attempt to match the generated data distribution to real data distribution, namely:

$$\mathcal{L}_G = \|X - G(X)\|_1 + \|G(X) - D(G(X))\|_1$$

Pixel-wise error ($\ell_1$) is chosen to achieve sharper results, following insights from Isola et al. [13]. Furthermore, since both sub-networks have similar capability, we can further improve the training stability with balancing generator and discriminator's learning rates proposed by Sainburg et al. [14]. The learning rates are modeled by a sigmoid centered around zero, allowing the weaker performing network to catch up at the expense of the stronger network (according to some relative performance indicator of the two networks). Each sub-network only consists of fully convolution layers with no fully connected layers. This effectively reduces the number of parameters to learn, results in a more robust model. Furthermore, (de)convolutional layers with strides are used, instead of pooling/upsampling layers which greatly helps with image reconstruction quality as proposed in DCGAN [15].

### 3.1 Anomaly score

We leverage the reconstruction error in the discriminator as the anomaly score, since the discriminator is trained to separate the normal and generated data distribution.

As the model converges, the generator would have learnt to reconstruct normal data that belong to original distribution $X$. Normal data passing through the generator would have no problem being reconstructed again at the discriminator. On the other hand, anomalous inputs would cause the generator to fail to reconstruct faithfully. The reconstruction error would then be amplified at the discriminator, having thought the inputs from the generator does not belong to the normal data distribution. Figure 2 illustrates the concept more clearly for the case of MNIST data with 0 as anomaly class.

Formally, for each input $\hat{x}$, anomaly score $\mathcal{A}(\hat{x})$ is calculated as:

$$\mathcal{A}(\hat{x}) = \|\hat{x} - D(G(\hat{x}))\|_2$$

This choice is in contrast with many other anomaly detection methods which uses generator reconstruction error for calculating anomaly scores. We notice that our choice of anomaly score calculation allows the better split between normal and anomalous scores, leading to superior performance. We then define a threshold $\phi$ from which anomalies are determined. A test input $\hat{x}_i$ is considered to be anomalous if $\mathcal{A}(\hat{x}_i) > \phi$.

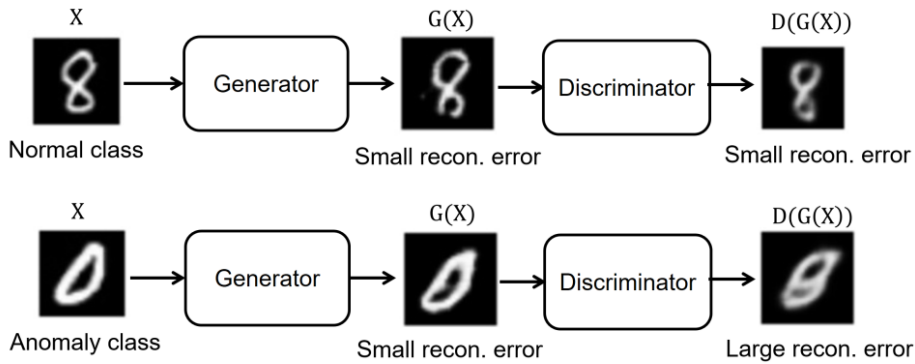

**Figure 2:** For normal class 8. The generator could reconstruct the inputs with little error, causing the discriminator to successfully reconstruct the number 8. In the case of anomalous input 0, even though generator could still reconstruct the image, the discriminator still determines that the input is not from normal data distribution and fails to reconstruct the image, giving rise to high anomaly score.

## 4 Experiment

### 4.1 Datasets

Firstly, we apply ADAE to two benchmark datasets MNIST [16] and CIFAR-10 [17] for a fair comparison with other state-of-the-art methods. Then, we demonstrate the practicality of our solution with a real life problem of brain tumor detection using BRATS 2017 [18] dataset after training on healthy brain MRI images from Human Connectome Project (HCP) [19] dataset.

**MNIST and CIFAR-10:** for each run, one label is left out to act as anomaly class. The model is trained on the remaining nine classes as normal data. In total, there are 10 sets of data for each dataset, each corresponds to one class being anomalous. Training sets consists of 80% of normal data while the test set occupies the rest of the normal data and all anomalous data, following the setup from Zenati et al. [4]. Images are normalized in range [-1,1] and resized to 32x32xcolorChanels.

**HCP and BRATS 2017:** We make use of 65 healthy subjects in HCP dataset for training data and the full BRATS 2017 dataset for testing. BRATS 2017 dataset contains 210 patients with high-grade glioblastomas (HGG) and 75 patients with low-grade glioblastomas (LGG). The images undergo intensity normalization with z-score normalization to reduce variations in intensity of different subjects. Each slice is zero-padded to make a square (if applicable), and resized to 32x32 for simplicity.

Training is done with batch size of 64, Adam optimizer with parameters $\alpha = 10^{-5}, \beta = 0.5$ over 100 epochs. The latent space dimension for each case is different given the varying complexity of each dataset and the known weakness of autoencoder. Too few dimensions will cause a vital information loss while too large dimensions will result in the model simply trying to replicate the inputs without filtering out redundant information. Details of architecture choices for each dataset are outlined in Appendix A.

### 4.2 Evaluation metrics

After every instance, with their ground truth labels, in the test set is given an anomaly score, Receiver Operating Characteristic (ROC) curve can be plotted with True Positive Rate (TPR) and False Positive Rate (FPR) pair at different thresholds $\phi$. Area under ROC (AUROC) can then be obtained to indicate the performance of the model on each dataset. ROC is chosen as they are unaffected by skewed datasets whereby anomalous cases are uncommon.

For brain MRI datasets, binary labels (normal or tumor) are decided based on the presence of tumor mask in ground truth annotations. Here we are not particularly focus on the tumor segmentation but only predicting whether an image contains tumor, also called slice-wise detection. Anomaly score, along with its ground truth label, for each test slice is obtained for AUROC score calculation.

## 5 Results

Table **1** shows the MNIST performance of ADAE against different methods such as IGMM-GAN [6], GANomaly [3], EGBAD [4] and AnoGAN [20]. ADAE consistently outperforms GANomaly, EGBAD and AnoGAN in most of the classes while having slightly better performance as compared to IGMM-GAN which specializes in multimodal datasets such as MNIST, average AUROC of 0.858 as compared to 0.852.

**Table 1:** AUC scores for each anomalous class on MNIST dataset

| Anomaly | 0 | 1 | 2 | 3 | 4 | 5 | 6 | 7 | 8 | 9 | Avg |
|---|---|---|---|---|---|---|---|---|---|---|---|
| AnoGAN | 0.610 | 0.300 | 0.535 | 0.440 | 0.430 | 0.420 | 0.475 | 0.355 | 0.400 | 0.335 | 0.430 |
| EGBAD | 0.775 | 0.290 | 0.670 | 0.520 | 0.450 | 0.430 | 0.570 | 0.400 | 0.545 | 0.345 | 0.500 |
| GANomaly | 0.880 | 0.650 | 0.940 | 0.780 | 0.870 | 0.840 | 0.830 | 0.660 | 0.830 | 0.520 | 0.780 |
| IGMM-GAN | **0.955** | **0.900** | 0.930 | 0.820 | **0.830** | 0.900 | **0.930** | **0.900** | 0.780 | 0.570 | 0.852 |
| ADAE (ours) | 0.951 | 0.821 | **0.948** | **0.889** | 0.819 | **0.906** | 0.889 | 0.803 | **0.925** | **0.631** | **0.858** |

Similarly, for a considerably more difficult dataset CIFAR-10, ADAE also outperforms all of the existing methods with average AUROC of 0.610.

**Table 2:** AUC scores for each anomalous class on CIFAR-10 dataset

| Anomaly | Plane | Car | Bird | Cat | Deer | Dog | Frog | Horse | Ship | Truck | Avg |
|---|---|---|---|---|---|---|---|---|---|---|---|
| AnoGAN | 0.516 | 0.492 | 0.411 | 0.399 | 0.335 | 0.393 | 0.321 | 0.399 | 0.567 | 0.511 | 0.434 |
| EGBAD | 0.577 | 0.514 | 0.383 | 0.448 | 0.374 | 0.481 | 0.353 | 0.526 | 0.413 | 0.555 | 0.462 |
| GANomaly | 0.625 | 0.629 | 0.505 | 0.577 | **0.593** | **0.633** | **0.653** | 0.601 | **0.622** | 0.614 | 0.605 |
| ADAE(ours) | **0.633** | **0.729** | **0.550** | **0.580** | 0.496 | 0.599 | 0.590 | **0.610** | 0.619 | **0.697** | **0.610** |

In brain MRI problem domain, specifically BRATS 2017 dataset, ADAE outperforms different common methods in slice-wise tumor detection such as ceVAE, VAE and AE with 0.892 AUC score, as shown in Figure 3.

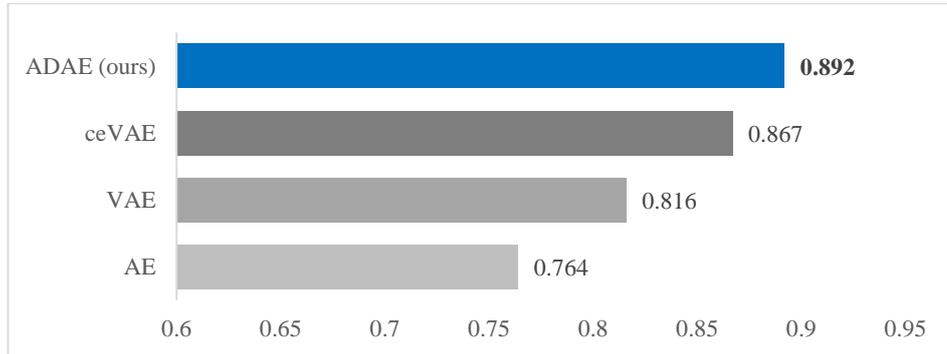

**Figure 3:** AUC scores for BRATS 2017 dataset

Figure 4 illustrates both qualitative and quantitative aspects of ADAE for BRATS 2017 dataset. Having learnt only normal data, the model would reconstruct the normal brain image from the test inputs, causing the images being different at the tumor position. This causes a good split between normal and anomaly class in the histogram, demonstrating ADAE's ability to detect brain tumor reliably.

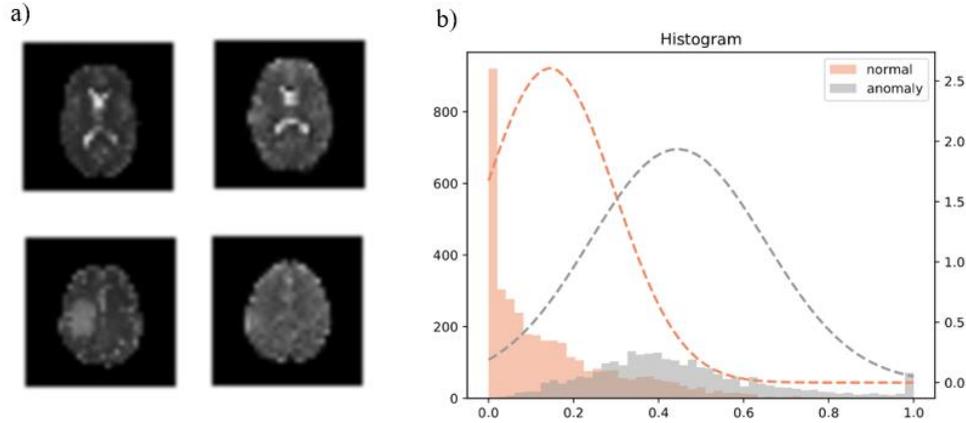

**Figure 4:** a) Reconstruction quality of generated image (right) from test inputs (left). Top row: normal data being reconstructed faithfully; bottom row: anomalous data being reconstructed without the tumor, having learnt the normal data. b) Histogram of anomaly scores assigned to each test input for both normal and anomaly class.

# 6  Conclusion

In this work, we introduced a GAN-based method for anomaly detection which uses two autoencoders as generator and discriminator respectively. We also made use of discriminator reconstruction error as anomaly score for better detection. The final results in both benchmark datasets MNIST and CIFAR-10 as well as real life use case in brain tumor detection showed that the model is robust in different problem domains.

# Appendix A

### a. MNIST experiment details

Latent space dimension: 32

| Component | Kernel x Strides | Features maps | Batch Norm. | Activation |
|---|---|---|---|---|
| G_Enc | 3x2 | 16 | | ReLU |
| | 3x2 | 32 | ✓ | ReLU |
| | 3x2 | 48 | ✓ | ReLU |
| | 4x1 | 16 | ✓ | ReLU |
| | 1x1 | 32 | ✓ | ReLU |
| G_Dec | 4x1 | 16 | ✓ | LeakyReLU |
| | 3x2 | 16 | ✓ | LeakyReLU |
| | 3x2 | 16 | ✓ | LeakyReLU |
| | 3x2 | 32 | | Tanh |
| D_Enc | 3x2 | 16 | | ReLU |
| | 3x2 | 32 | ✓ | ReLU |
| | 3x2 | 48 | ✓ | ReLU |
| | 4x1 | 16 | ✓ | ReLU |
| | 1x1 | 32 | ✓ | ReLU |
| D_Dec | 4x1 | 16 | ✓ | LeakyReLU |
| | 3x2 | 16 | ✓ | LeakyReLU |
| | 3x2 | 16 | ✓ | LeakyReLU |
| | 3x2 | 32 | | Tanh |

## b. CIFAR-10, HCP and BRATS 2017 experiment details

Latent space dimension: 128. These datasets have considerably more details than MNIST, requiring more latent dimension to retain useful features in the latent space for reconstruction.

| Component | Kernel x Strides | Features maps | Batch Norm. | Activation |
|---|---|---|---|---|
| G_Enc | 3x2 | 64 |  | ReLU |
|  | 3x2 | 128 | ✓ | ReLU |
|  | 3x2 | 192 | ✓ | ReLU |
|  | 4x1 | 64 | ✓ | ReLU |
|  | 1x1 | 128 | ✓ | ReLU |
| G_Dec | 4x1 | 64 | ✓ | LeakyReLU |
|  | 3x2 | 64 | ✓ | LeakyReLU |
|  | 3x2 | 64 | ✓ | LeakyReLU |
|  | 3x2 | 32 |  | Tanh |
| D_Enc | 3x2 | 64 |  | ReLU |
|  | 3x2 | 128 | ✓ | ReLU |
|  | 3x2 | 192 | ✓ | ReLU |
|  | 4x1 | 64 | ✓ | ReLU |
|  | 1x1 | 128 | ✓ | ReLU |
| D_Dec | 4x1 | 64 | ✓ | LeakyReLU |
|  | 3x2 | 64 | ✓ | LeakyReLU |
|  | 3x2 | 64 | ✓ | LeakyReLU |
|  | 3x2 | 32 |  | Tanh |